\documentclass[letterpaper, 10 pt, conference]{ieeeconf}  

\IEEEoverridecommandlockouts                              

\title{\LARGE \bf
An Improved Wrist Kinematic Model for Human-Robot Interaction
}

\author{Ningbo Yu$^{1}$ and Chang Xu$^{2}$
\thanks{This work was supported by 
		the National Natural Science Foundation of China (61720106012, 61403215) and
		the Fundamental Research Funds for the Central Universities.}
\thanks{$^{1}$Corresponding author Assoc. Prof. Dr. Ningbo Yu is with the Institute of Robotics and Automatic Information Systems,
	Nankai University, Haihe Education Park, Tianjin 300353, China. Email:
	{\tt\small  nyu@nankai.edu.cn}}%
\thanks{$^{2}$Chang Xu is with the Department of Systems and Information Engineering, University of Virginia,
	Charlottesville, VA 22903, USA. Email:
	{\tt\small cx5dq@virginia.edu}}%
}

\usepackage[american]{babel}
\usepackage{microtype}
\usepackage{graphicx}
\DeclareGraphicsExtensions{.pdf,.ps,.eps}

\begin{document}

\maketitle
\thispagestyle{empty}
\pagestyle{empty}

\begin{abstract}

Human kinematics is of fundamental importance for rehabilitation and assistive robotic systems that physically interact with human. The wrist plays an essential role for dexterous human-robot interaction, but its conventional kinematic model is oversimplified with intrinsic inaccuracies and its biomechanical model is too complicated for robotic applications. In this work, we establish an improved kinematic model of the wrist. \emph{In vivo} kinematic behavior of the wrist was investigated through noninvasive marker-less optical tracking. Data analysis demonstrated the existence of measurable dynamic axes in carpal rotation, justifying inevitable misalignment between the wrist and robotic representation if using the conventional wrist model. A novel wrist kinematic model was then proposed with rigid body transformation in fusion with a varying prismatic
term indicating the dynamic axes location. Accurate and real-time estimation of this term has been achieved through coupled
wrist angles with nonlinear regression. The proposed model is not only accurate but also conveniently applicable for translating anatomical behaviors into robotic implementation and functional assessment for precise and dexterous human-robot interaction.

\end{abstract}

\section{INTRODUCTION}

Neurorehabilitation therapies promote partial or complete restore upper extremity functions after injury to the central or peripheral nervous system~\cite{klamroth2014three,metzger2014assessment}. Compared with conventional constrained-induced therapeutic approaches, robot-assisted therapies have demonstrated potential advantages on adaptive engagement, quantitative evaluation of rehabilitation process, and clinical assessment~\cite{maciejasz2014survey,atashzar2016characterization,xu2017quantitative}. Moreover, for physical human-robot interaction in rehabilitation and assistive applications, human kinematics is of fundamental importance.

Motion tracking and kinematic analysis come all the way with robotic neurorehabilitation treatment and human-robot interaction~\cite{rinderknecht2018reliability,nef2009armin}. Evaluation of upper extremity kinematics in both Cartesian and joint space and ergonomic design of robotic exoskeletons are fundamentally based on accurate skeletal modeling~\cite{santos2006reported,wang2016extended}. In specific, modeling and quantification of wrist kinematics are challenging due to the articulation of eight carpal bones and anatomical variability within carpus. Conventional models approximate the skeletal structure as universal joints linking rigid segments of the upper limb~\cite{maciejasz2014survey,babaiasl2016review}. However, undesired misalignment between robotic device and human joints and discrepancies in resolving parametric kinematics inevitably occur due to the assumption that there is only rotational motion in wrist joints and the upper limb is decoupled into independent segments~\cite{nef2009armin}. Sophisticated coupled mechanism and dexterous movements of the upper extremity can not be fully investigated and explained through the oversimplified conventional kinematic model. Therefore, functional assessments on wrist and mechanical implementation of rehabilitation devices would have deficits in kinematic analysis and biomechanical evaluation.

Different from conventional kinematic models, some studies have investigated the coupled limb biomechanics quantitatively. The coupled arm impedance measured by dynamic characteristics like inertia, damping and stiffness were presented and estimated under muscular co-contraction~\cite{patel2014effect}. Flexible movements in distal arm are essential for dexterous manipulation and passive stiffness of those coupled joints were also estimated quantitatively~\cite{formica2012passive}. 

Besides robotic rehabilitation studies, biomechanical investigations of wrist and forearm provide detailed descriptive data and rational estimation of joint kinematics. An improved elasto-kinematic model of the forearm was proposed and it justified the relative motion (displacements between forearm bones) in proximal radio-ulnar joint instead of describing radio-ulnar joints as universal joints~\cite{kecskemethy2005improved}. Both \emph{in vitro} and \emph{in vivo} studies on carpal bone kinematics during wrist motion demonstrated that there is no fixed instantaneous screw axes of wrist rotation and sophisticated carpal joints can not be simplified as universal joints due to the evident translation between carpal bones~\cite{patterson1998high,neu2001vivo}. To assess kinematics and dynamics of wrist joints, rotational behavior of primary carpal bones was described quantitatively and more realistic coordinate systems were also defined~\cite{foumani2009vivo,coburn2007coordinate}. 

However, anatomical characteristics and kinematic behavior presented in biomechanical studies are dedicated to individual carpal bones and specialized coordinate systems are not applicable in functional assessment for robotic neurorehabilitation. Human-robot interaction involves global movement of end-effectors and distal joints. Those biomechanical coordinate systems are too sophisticated that kinematic analysis intended for robotic evaluation is impracticable to conclude from distinct individual behavior of various carpal bones.

To address aforementioned oversimplification and this inapplicability situation, based on anatomical investigation and biomechanical behavior of coupled distal arm joints, the purpose of this study is to establish a more accurate kinematic model of the wrist which is more applicable for translating anatomical behaviors into robotic implementation and functional assessment from precise dexterous human-robot interaction. \emph{In vivo} kinematic behavior of carpal joint during wrist flexion-extension~(FE) was investigated through noninvasive markerless optical tracking method. Experiments with 25 healthy uninjured subjects have validated the feasibility and precision of this improved kinematic model and provide the justification of translational rotation axes through nonlinear regression quantitatively.

\section{Methods}

\subsection{Conventional Model of the Upper Limb}

Given the assumptions that the skeleton of human body is comprised of rigid bone segments which are linked by frictionless and universal joints, conventional kinematic model of the upper limb were introduced to approximate the movement of human body in terms of ranges of motion~(ROMs) and degrees of freedom~(DOFs).

\begin{figure}[b]
	\centering	
	\includegraphics[height=0.65\columnwidth]{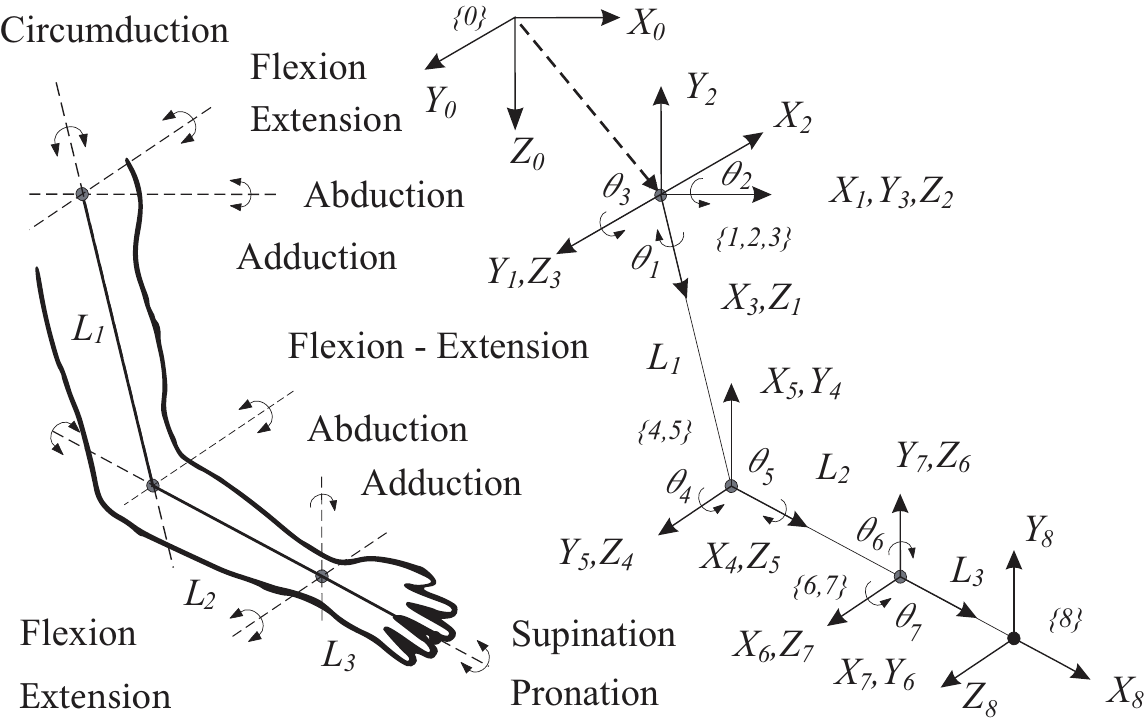}
	\caption{Conventional model of the upper limb~\cite{babaiasl2016review}. The right upper limb is modeled as three rigid segments linked by three universal joints which are the shoulder, the elbow and the wrist joint, denoted as $\left\{ 1 \right\}$, $\left\{ 4 \right\}$ and $\left\{ 6 \right\}$ respectively. Position and orientation of the end-effector is described in coordinate frame $\left\{ 8 \right\}$ with respect to base coordinate frame $\left\{ 0 \right\}$. 7 DOFs movements of the upper limb are described through rotation angles ${\theta _i}$ between consecutive frames.}
	\label{conventional_model}
\end{figure}

The upper limb includes three rigid segments which are anatomically described as the upper arm, the forearm and the hand (considered as a single segment) respectively. Three universal joints link these segments which are the shoulder, the elbow and the wrist joint~\cite{babaiasl2016review}. With the assumptions that only rotational motion occurs in these joints, mechanical equilibrium equations can be established and conventional kinematic model for upper limb with a total of 7 DOFs is proposed, as shown in Fig.~\ref{conventional_model}. The base of conventional coordinate system $\left\{ 0 \right\}$ is located in the midway between shoulders and three frames are located at the center of the complex joint shoulder. The human wrist is considered as a universal joint with 2 DOFs: abduction-adduction~(radio-ulnar deviation, RUD)~(${X_6}{Y_6}{Z_6}$) and FE~(${X_7}{Y_7}{Z_7}$). The end-effector frame $\left\{ 8 \right\}$ is located at the extended fingertips as illustrated in Fig.~\ref{conventional_model}. Therefore, the position and orientation of the end-effector~(human hand) can be calculated from forward kinematics and the angular movement in joint space can be approximated through inverse kinematics.

\subsection{Carpal Kinematic Behavior in Wrist Motion}

\begin{figure}[h]
	\centering	
	\includegraphics[height=0.9\columnwidth]{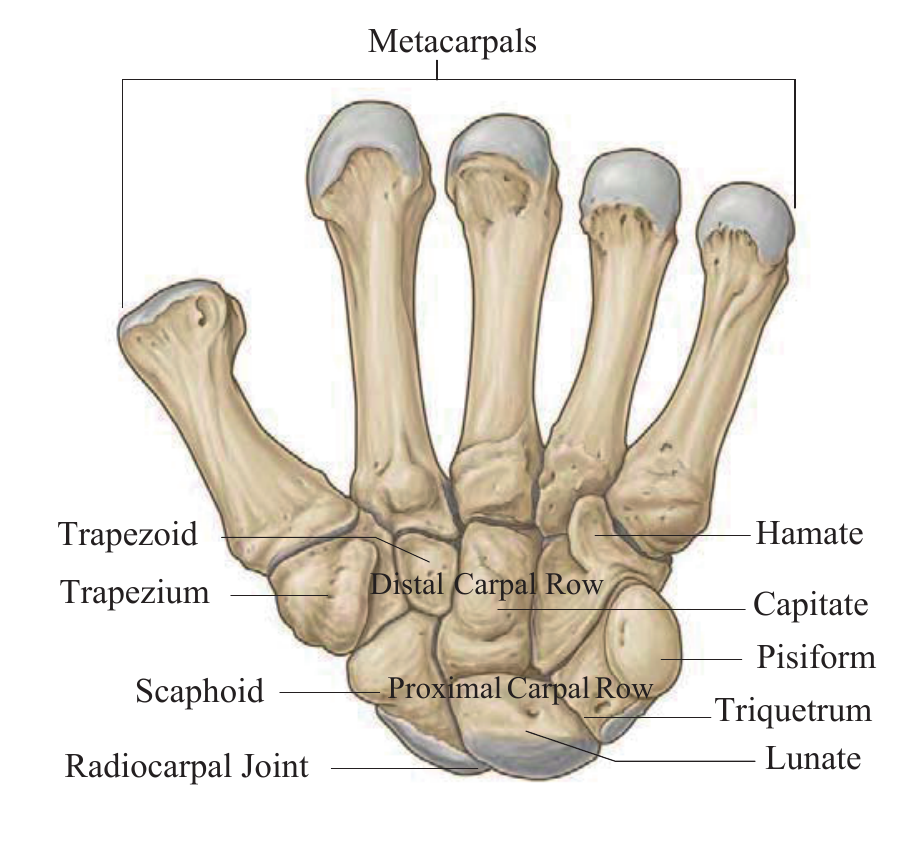}
	\caption{Volar view of the carpal and metacarpal bones of the left hand in functional neutral wrist position~\cite{standring2015gray}. With consistent kinematic behavior constrained by carpal row alignment, wrist motion occurs as the combination movement of radiocarpal and midcarpal joints with unfixed rotation axes within the capitate.}
	\label{carpal_bones}
\end{figure}

The carpus region is comprised of eight dexterous and intricately shaped carpal bones which articulate with each other complexly and interplay with the metacarpal bones and the distal radio-ulnar joint. From volar view of a left wrist, as shown in Fig.~\ref{carpal_bones}, those carpal bones are arranged into the proximal row and the distal row. From the anterior view of the proximal row, scaphoid, lunate, triquetrum and pisiform are arranged from lateral to medial. The trapezium, trapezoid, capitate and hamate make up the distal row accordingly from lateral to medial~\cite{drake2014gray}. 

Constrained by carpal ligaments and combined with distinct geometric articulation, each carpal bone has six DOFs, enabling a virtually hemisphere of the wrist motion and the complexity of carpal kinematics. Therefore, it is difficult to describe the complete wrist motion precisely based on individual carpal bone kinematic behavior~\cite{coburn2007coordinate}. According to the anatomical basis of clinical practice, radiocarpal joint, intercarpal joint (including joint of proximal row, midcarpal joint and joint of distal row) and carpometacarpal joint comprise the overall wrist joints~\cite{standring2015gray}. Previous studies have explained the rationale of simplifying the wrist motion into movement at radiocarpal and midcarpal joints, based on the segments of proximal and distal carpal rows which are regarded as rigid and tight structure~\cite{drake2014gray,wolfe2000vivo}. The radiocarpal joint is a typical condyloid joint with two axes formed between the proximal row and the distal end of the radius, allowing biaxial movement of FE, abduction-adduction and circumduction. Carpal bones within the same carpal row which are reinforced by ligaments exhibit consistent kinematic behavior, resulting virtually no relative movement at both proximal and distal joint~\cite{wolfe2000vivo}. Therefore, the movement at intercarpal joints is together with radiocarpal actuated by the same muscle, and is limited in midcarpal joint. Based on anatomical characteristics and rational simplified segments of the carpus, the sophisticated wrist motion can be described as FE in the frontal plane, RUD in the sagittal plane and circumduction, which mostly occur in the combination movement of radiocarpal and midcarpal joints.

Furthermore, the capitate is referred as the keystone of carpus in distal row, and there is relatively minimal movement between capitate and the third metacarpal distally. The capitate also exhibits articular engagement with the scaphoid and lunate proximally~\cite{wolfe2000vivo}. From both clinical function and biomechanics point of view, the capitate indicates wrist motion due to its prominent position and it is reasonable to locate rotation axes within the capitate during wrist FE and RUD.

According to the conventional model, the coordinate frame is modeled as a universal joint with 2 DOFs. Only rotational motion about the pivot point can be described while linear displacement along the axes is excluded from the conventional kinematic model. However, both \emph{in vitro} and \emph{in vivo} studies on carpal bone kinematics during wrist motion demonstrated that the center of carpal rotation is not fixed in the proximal pole of the capitate. Variations in the location of pivot point and evident measurable translation in carpal bones were reported~\cite{patterson1998high,neu2001vivo}. The kinematics of FE and RUD of the human wrist is much more sophisticated and exhibits distinct carpal mechanisms, rather than being modeled as a simplified fixed hinge or simple joint as indicated by conventional model of the upper limb.

\subsection{Improved Kinematic Model of Wrist Motion}

\begin{figure}[!h]
	\centering	
	\includegraphics[height=1.2\columnwidth]{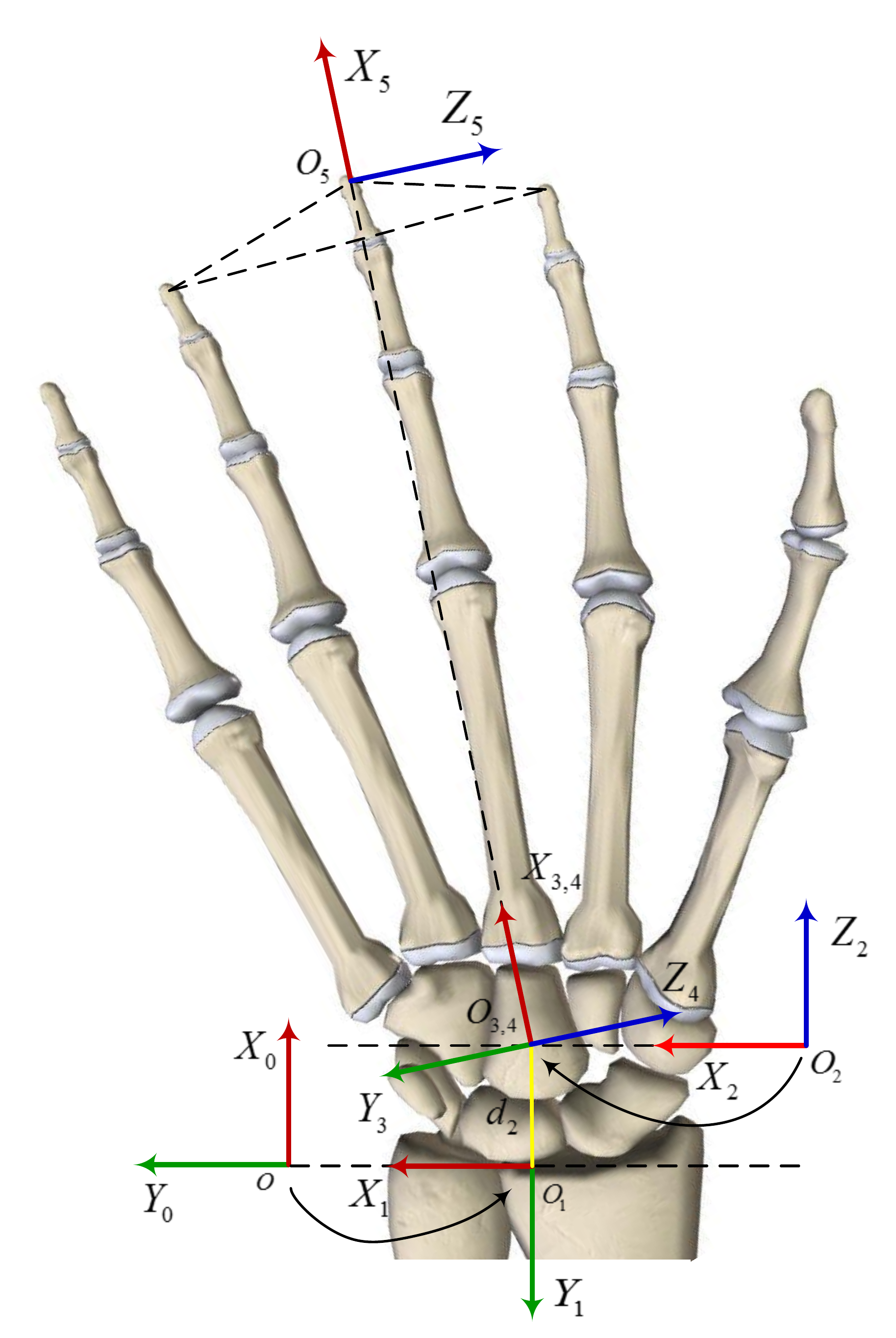}
	\caption{Improved kinematic model of wrist motion is illustrated in volar view of right wrist. With respect to base coordinate system $\left\{ 0 \right\}$, position and orientation of the end-effector is described in coordinate frame $\left\{ 5 \right\}$ which is located within the plane defined by three fingertips. 2 DOFs movements of the wrist are described through rotation angles in coordinate frame $\left\{ 3 \right\}$ and $\left\{ 4 \right\}$ which indicate angles in RUD and FE respectively. Translation parameter ${d_2}$ indicates the variable location of rotation center for global wrist motion.}
	\label{improved_model}
\end{figure}

Compared with conventional model of the upper limb (see Fig.~\ref{conventional_model}) which illustrates that the carpal rotation occurs about a single pivot point and wrist joints act like universal joints, the proposed improved model (see Fig.~\ref{improved_model}) introduces a prismatic joint to incorporate the 2 DOFs revolute joints, locates the rotation center and describes the translational behavior of the axes in a more precise manner. It presents a biomechanical rational description of the carpal behavior and more accurate estimation of the position and orientation of the end-effector~(fingertip) for wrist kinematics and upper limb movement.

\begin{table}
	\caption{D-H Parameters of the improved model}
	{\renewcommand{\arraystretch}{1.8}
		{\begin{tabular}{p{1cm} p{1cm} p{1cm} p{1cm} p{1cm} p{1cm}}
				\hline
				\hline
				$i$ & ${\alpha _{i - 1}}$ & ${a_{i - 1}}$ & ${d_i}$ & ${\theta _i}$ & ${\beta _i}$ \\
				\hline
				1 & 0 & 0 & 0 & ${90^ \circ }$ \\
				
				2 & ${90^ \circ }$ & 0 & ${d_2}$ & 0 \\
				
				3 & ${-90^ \circ }$ & 0 & 0 & ${\theta _3}$ & ${\beta _3} - {90^ \circ }$\\
				
				4 & ${90^ \circ }$ & 0 & 0 & ${\theta _4}$ & ${\beta _4}$ \\
				
				5 & 0 & ${a_4}$ & 0 & 0 \\
				\hline
				\hline
		\end{tabular}}
		\label{improved_DH}
		\par}
\end{table}

According to Denavit-Hartenberg~(D-H) convention \cite{denavit1955kinematic}, the origin of improved coordinate system locates within the radiocarpal joint, at the proximal articular lunate surface for radius (as in Fig.~\ref{improved_model}). The positive ${X_0}$ axis parallels to the radial long axis and points distally. The positive ${Y_0}$ axis is directed ulnarly through the radial styloid and is perpendicular to the ${X_0}$ axis. The coordinate system $\left\{ 1 \right\}$ shares the same origin with $\left\{ 0 \right\}$ and rotates about ${Z_1}$ axis for ${90^ \circ }$ from $\left\{ 0 \right\}$, according to the right-hand rule. The consistent origins of the coordinate system $\left\{ 2 \right\}$, $\left\{ 3 \right\}$ and $\left\{ 4 \right\}$ are located within the capitate in initial configuration (no movement in FE or RUD), as the indicator of the center of global wrist motion. Due to the fact that the relative motion between the third metacarpal and capitate is clinically negligible and the mounting fixture rigidly constrains the middle finger with the palm (as in Fig.~\ref{conventional_model}), the ${X_{3,4,5}}$ and ${Z_2}$ axes coincide with the straight line from ${O_2}$ to middle fingertip ${O_5}$ distally. The translation along the ${Z_2}$ axis is denoted as ${d_2}$, representing the variation in the location of rotation center for global wrist motion. The positive ${X_2}$ and ${Y_3}$ axes coincide with ${Y_0}$ initially. The positive ${Z_3}$ axis represents the rotation axis for RUD volarly. The angle ${\beta _3}$ of RUD (positive in ulnar deviation) derives from rotation ${\theta _3}$ around ${Z_3}$ axis (as in Table~\ref{improved_DH}). The coordinate system $\left\{ 4 \right\}$ rotates about ${X_3}$ axis for ${90^ \circ }$ from $\left\{ 3 \right\}$. The positive ${Z_4}$ axis indicates the rotation axis for FE radially and the angle ${\theta _4}$ equals to ${\beta _4}$ which is positive in flexion. The coordinate system $\left\{ 5 \right\}$ is located within the plane defined by index, middle and ring fingertips which is coincident with volar plane due to the constraint fixture. It represents the position and orientation of the end-effector.

\subsection{Kinematics Analysis of the Improved Model}

\subsubsection{Forward kinematics}

The improved kinematic model introduces a prismatic joint (as joint 2 in Fig.~\ref{improved_model}) to incorporate the revolute joints (as joint 3,4 in Fig.~\ref{improved_model}) and describes the carpal behavior during wrist motion based on biomechanical rationale. The general transformation matrix for coordinate system $\left\{ i \right\}$ with respect to $\left\{ {i-1} \right\}$ is given by
\begin{equation}\label{gen_trans}
\begin{array}{l}
{}_i^{i - 1}T = Ro{t_X}\left( {{\alpha _{i - 1}}} \right)Tran{s_X}\left( {{a_{i - 1}}} \right)Ro{t_Z}\left( {{\theta _i}} \right)Tran{s_Z}\left( {{d_i}} \right)\\
\;\;{\kern 1pt} {\kern 1pt} {\kern 1pt} {\kern 1pt} {\kern 1pt} {\kern 1pt} {\kern 1pt} {\kern 1pt} {\kern 1pt} {\kern 1pt} {\kern 1pt} {\kern 1pt}  = \left[ {\begin{array}{*{20}{c}}
	{c{\theta _i}}&{ - s{\theta _i}}&0&{{a_{i - 1}}}\\
	{s{\theta _i}c{\alpha _{i - 1}}}&{c{\theta _i}c{\alpha _{i - 1}}}&{ - s{\alpha _{i - 1}}}&{ - s{\alpha _{i - 1}}{d_i}}\\
	{s{\theta _i}s{\alpha _{i - 1}}}&{c{\theta _i}s{\alpha _{i - 1}}}&{c{\alpha _{i - 1}}}&{c{\alpha _{i - 1}}{d_i}}\\
	0&0&0&1
	\end{array}} \right]
\end{array}\\\,
\end{equation}
where ${c{\theta _i}}$ and ${s{\theta _i}}$ represent $\cos {\theta _i}$ and $\sin {\theta _i}$ respectively, $Ro{t_Z}\left( \Theta  \right)$ and $Tran{s_Q}\left( q \right)$ refer to rotational operators and translational operators respectively. The forward kinematics of the improved model derives from the general homogeneous transformation and the D-H parameters. The transformation matrices between two successive joints are calculated as:
\begin{equation}\label{sub_trans}
\begin{array}{l}
{}_1^0T = \left[ {\begin{array}{*{20}{c}}
	0&{ - 1}&0&0\\
	1&0&0&0\\
	0&0&1&0\\
	0&0&0&1
	\end{array}} \right]{\kern 1pt} {\kern 1pt} {\kern 1pt} {\kern 1pt} {\kern 1pt} {\kern 1pt} {\kern 1pt} {\kern 1pt} {\kern 1pt} {\kern 1pt} {\kern 1pt} {\kern 1pt} {\kern 1pt} {\kern 1pt} {\kern 1pt} {\kern 1pt} {\kern 1pt} {\kern 1pt} {\kern 1pt} {\kern 1pt} {\kern 1pt} _2^1T = \left[ {\begin{array}{*{20}{c}}
	1&0&0&0\\
	0&0&{ - 1}&{ - {d_2}}\\
	0&1&0&0\\
	0&0&0&1
	\end{array}} \right]\\
\\
_3^2T = \left[ {\begin{array}{*{20}{c}}
	{{c_3}}&{ - {s_3}}&0&0\\
	0&0&1&0\\
	{ - {s_3}}&{ - {c_3}}&0&0\\
	0&0&0&1
	\end{array}} \right]{\kern 1pt} {\kern 1pt} {\kern 1pt} _4^3T = \left[ {\begin{array}{*{20}{c}}
	{{c_4}}&{ - {s_4}}&0&0\\
	0&0&{ - 1}&0\\
	{{s_4}}&{{c_4}}&0&0\\
	0&0&0&1
	\end{array}} \right]\\
\\
_5^4T = \left[ {\begin{array}{*{20}{c}}
	1&0&0&{{a_4}}\\
	0&1&0&0\\
	0&0&1&0\\
	0&0&0&1
	\end{array}} \right]
\end{array}
\end{equation}
To determine the posture of the end-effector in the improved model, the forward kinematics with respect to the origin coordinate system $\left\{ 0 \right\}$ is calculated as:
\begin{equation}\label{05_trans_1}
\begin{array}{l}
{}_5^0T = {}_1^0T{}_2^1T{}_3^2T{}_4^3T{}_5^4T\\
\\
{\kern 1pt} {\kern 1pt} {\kern 1pt} {\kern 1pt} {\kern 1pt} {\kern 1pt} {\kern 1pt} {\kern 1pt} {\kern 1pt} {\kern 1pt} {\kern 1pt} {\kern 1pt}  = \left[ {\begin{array}{*{20}{c}}
	{ - {s_3}{c_4}}&{{s_3}{s_4}}&{{c_3}}&{{d_2} - {a_4}{s_3}{c_4}}\\
	{{c_3}{c_4}}&{ - {c_3}{s_4}}&{{s_3}}&{{a_4}{c_3}{c_4}}\\
	{{s_4}}&{{c_4}}&0&{{a_4}{s_4}}\\
	0&0&0&1
	\end{array}} \right]\,.
\end{array}
\end{equation}

\subsubsection{Inverse kinematics}

\begin{figure}[h]
	\centering	
	\includegraphics[height=0.55\columnwidth]{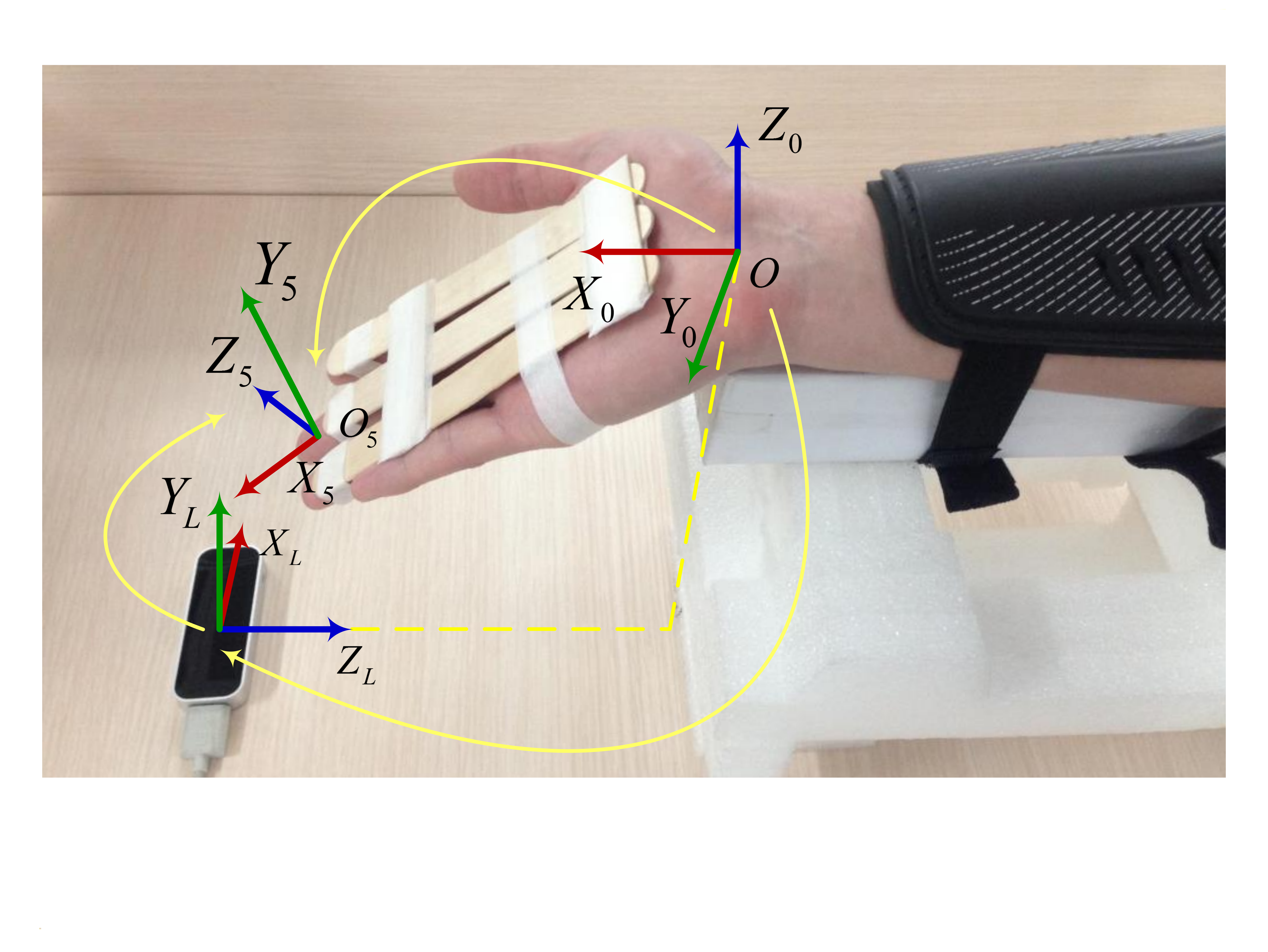}
	\caption{Position and orientation of coordinate frame $\left\{ 5 \right\}$ with respect to $\left\{ L \right\}$ are recorded via Leap Motion. The homogeneous transformation matrix linking $\left\{ L \right\}$ and $\left\{ 0 \right\}$ is derived from experimental configuration.}
	\label{transform}
\end{figure}

According to the posture of the end-effector detected through optical sensor in Cartesian space, the inverse kinematics determines the rotation angle and axes location in joint space during carpal motion. As illustrated in Fig.~\ref{transform}, the sensor detects the discrete position and orientation of the middle fingertip in the coordinate system $\left\{ L \right\}$. The ${X_L}$ axis lies horizontally and parallels to the edge of the device and ${Y_0}$ axis in the wrist. The positive ${Y_L}$ axis is perpendicular to the plane ${X_0}{Y_0}$ and is directed volarly.

The posture data of the end-effector should be transformed into the improved model for inverse kinematics calculation. Given the optical sensor coordinate system, the origin coordinate frame and the rotation matrix, denoted as $\left\{ L \right\}$, $\left\{ 0 \right\}$ and $R\left( \theta  \right)$ respectively, the homogeneous transformation matrix which links the two coordinate system is described by
\begin{equation}\label{0l_trans}
\begin{array}{l}
{}_L^0T = \left[ {\begin{array}{*{20}{c}}
	{{}_L^0R\left( \theta  \right)}&{{}^0{P_{LORG}}}\\
	0&1
	\end{array}} \right] {\kern 1pt}
{}_L^0R\left( \theta  \right) = \left[ {\begin{array}{*{20}{c}}
	0&0&{ - 1}\\
	{ - 1}&0&0\\
	0&1&0
	\end{array}} \right]\
\end{array}
\end{equation}
where ${{}^0{P_{LORG}}}$ represents the location of origin ${O_L}$ with respect to the coordinate system $\left\{ 0 \right\}$ which is measured according to the experimental configuration. The discrete position and orientation of the end-effector $\left\{ 5 \right\}$ tracked by the optical sensor are described with respect to the coordinate system $\left\{ L \right\}$ and denoted as ${}_5^LT$. Homogeneous transformation is applied for the detected posture with respect to the origin coordinate system $\left\{ 0 \right\}$:
\begin{equation}\label{05_trans_2}
\begin{array}{l}
{}_5^0T = {}_L^0T{}_5^LT\\
\\
{\kern 1pt}{\kern 1pt} {\kern 1pt} {\kern 1pt} {\kern 1pt} {\kern 1pt} {\kern 1pt} {\kern 1pt} {\kern 1pt} {\kern 1pt} {\kern 1pt} {\kern 1pt}  = \left[ {\begin{array}{*{20}{c}}
	{{n_x}}&{{o_x}}&{{a_x}}&{{p_x}}\\
	{{n_y}}&{{o_y}}&{{a_y}}&{{p_y}}\\
	{{n_z}}&{{o_z}}&{{a_z}}&{{p_z}}\\
	0&0&0&1
	\end{array}} \right]\
\end{array}
\end{equation}
According to the results derived from forward kinematics described in~(\ref{05_trans_1}), the angle ${\theta _4}$ in FE, ${\theta _3}$, ${\beta _3}$ in RUD and the translation ${d_2}$ can be calculated from the tracked data:
\begin{equation}\label{theta_4}
{\theta _4} = {\beta _4} = \arcsin \left( {{{{p_z}} \mathord{\left/
			{\vphantom {{{p_z}} {{a_4}}}} \right.
			\kern-\nulldelimiterspace} {{a_4}}}} \right)
\end{equation}
\begin{equation}\label{theta_3}
{\theta _3} = \arcsin \left( {{a_y}} \right), {\kern 1pt} {\kern 1pt} {\kern 1pt} {\beta _3} = {\theta _3} + \frac{\pi }{2}
\end{equation}
\begin{equation}\label{d_2}
{d_2} = {p_x} - {a_4}{n_x}\
\end{equation}
Therefore, the wrist rotation angle in FE and RUD with respect to neutral position can be estimated based on the posture of upper extremity. The translation parameter ${d_2}$ indicates the variation in carpal axes and justifies that the wrist joint can not be modeled as a simplified universal joint due to nonnegligible linear translation.

\subsubsection{Coupled joints fitting}

The evident translation along carpal axis is closely related to coupled carpal rotation. Previous study has demonstrated that statistically consistent carpal kinematic behavior can be found from different subjects~\cite{wolfe2000vivo}. Therefore, a model parameter fitting process was applied to find accurate numerical equation between wrist rotation center~(described by ${d_2}$ in prismatic joint) and coupled carpal angular motion~(${\beta _3}$ in RUD and ${\beta _4}$ in FE). A nonlinear regression model was established to fit observed data set points~$\left( {{x_i},{y_{\rm{i}}},{z_i}} \right)$, where~$\left( {x,y,z} \right)$ represented~$\left( {{\beta _3},{\beta _4},{d_2}} \right)$ and $i = 1,2, \ldots ,N$, to a nonlinear function which was a binary quadric polynomial described as
\begin{equation}\label{fit_z}
\hat z = \frac{{{a_1} + {a_3}x + {a_5}y + {a_7}{x^2} + {a_9}{y^2} + {a_{11}}xy}}{{1 + {a_2}x + {a_4}y + {a_6}{x^2} + {a_8}{y^2} + {a_{10}}xy}}
\end{equation}
where unknown parameters~${a_j}$ ($j = 1, \ldots ,10$) were optimized through Simple Genetic Algorithm~(SGA) to minimize the sum of square error~(SSE) between observation and estimation data sets
\begin{equation}\label{fit_sse}
SSE = \sum\limits_{i = 1}^N {\varepsilon _i^2}  = \sum\limits_{i = 1}^N {\left[ {{w_i}{{({z_i} - {{\hat z}_i})}^2}} \right]}\
\end{equation}
where ${\varepsilon _i}$ was residual term and ${w_i}$ was the weight set to be 1. Possible solutions of ${a_j}$ were encoded into the chromosome through vectors of real number coding and initial population size was set to be 20 empirically. Parent chromosome were chosen to mate through random selection and the crossover operators were developed by method of uniform crossover and the chance of crossover was set to be 0.85 empirically. Generally the rate of mutation was set to be 0.005 which was relatively low in consideration of the convergence rate and computational costs for the overall algorithm performance.

Measures of goodness of fit were applied to describe the fitness of the estimated parameters ${a_j}$ with the observation. The standardized residual ${r_i}$ derives from ${\varepsilon _i}$ and average residual $\bar \varepsilon $
\begin{equation}\label{fit_sr}
{r_i} = \frac{{{\varepsilon _i}}}{{\sqrt {\frac{1}{{N - 1}}\sum\limits_{i = 1}^N {{{({\varepsilon _i} - \bar \varepsilon )}^2}} } }}\
\end{equation}
The root mean square error (RMSE) indicates the deviation between estimated and observed values and is generally considered as a scale dependent residual metric
\begin{equation}\label{fit_rmse}
\begin{array}{l}
RMSE = \sqrt {MSE}  = \sqrt {{{SSE} \mathord{\left/
			{\vphantom {{SSE} N}} \right.
			\kern-\nulldelimiterspace} N}} \\
\\
{\kern 1pt} {\kern 1pt} {\kern 1pt} {\kern 1pt} {\kern 1pt} {\kern 1pt} {\kern 1pt} {\kern 1pt} {\kern 1pt} {\kern 1pt} {\kern 1pt} {\kern 1pt} {\kern 1pt} {\kern 1pt} {\kern 1pt} {\kern 1pt} {\kern 1pt} {\kern 1pt} {\kern 1pt} {\kern 1pt} {\kern 1pt} {\kern 1pt} {\kern 1pt} {\kern 1pt} {\kern 1pt} {\kern 1pt} {\kern 1pt} {\kern 1pt} {\kern 1pt} {\kern 1pt} {\kern 1pt} {\kern 1pt}  = \sqrt {\frac{1}{N}\sum\limits_{i = 1}^N {\left[ {{w_i}{{({z_i} - {{\hat z}_i})}^2}} \right]} }\
\end{array}
\end{equation}
The coefficient of determination, denoted as ${R^2}$, indicates the goodness of estimation in reflecting the variation of observations and is calculated as the square of the correlation between estimation and observation
\begin{equation}\label{fit_r_square}
\begin{array}{l}
{R^2} = \frac{{SSR}}{{SST}} = 1 - \frac{{SSE}}{{SST}}\\
\\
{\kern 1pt} {\kern 1pt} {\kern 1pt} {\kern 1pt} {\kern 1pt} {\kern 1pt} {\kern 1pt} {\kern 1pt} {\kern 1pt} {\kern 1pt} {\kern 1pt} {\kern 1pt} {\kern 1pt}  = \frac{{\sum\limits_{i = 1}^N {\left[ {{w_i}{{({{\hat z}_i} - {{\bar z}_i})}^2}} \right]} }}{{\sum\limits_{i = 1}^N {\left[ {{w_i}{{({z_i} - {{\bar z}_i})}^2}} \right]} }}{\kern 1pt} {\kern 1pt} {\kern 1pt} {\kern 1pt} {\kern 1pt} {\kern 1pt}  = 1 - \frac{{\sum\limits_{i = 1}^N {\left[ {{w_i}{{({z_i} - {{\hat z}_i})}^2}} \right]} }}{{\sum\limits_{i = 1}^N {\left[ {{w_i}{{({z_i} - {{\bar z}_i})}^2}} \right]} }}{\kern 1pt} {\kern 1pt} {\kern 1pt}\
\end{array}
\end{equation}
SSR represents the regression sum of squares. SST represents total sum of squares which is proportional to the variance of the data. The ${R^2}$ ranges from 0 to 1 statistically and ${R^2}$ of 1 indicates that the estimation fits the observation perfectly.

\subsection{Participants and Experimental Procedure}

Experiments have been designed and conducted to track human upper extremity movement and validate the proposed improved kinematic model. After obtaining informed consent, 25 healthy young subjects aged from 21 to 27~(4 females, age $22.8 \pm 1.3$ and 21 males, age $24.6 \pm 1.2$) were recruited to participate in the experiment. All subjects show no radiographic or history evidence of upper extremity pathology or chronic disease that might affect motion of distal arm. The movement of right wrist of each subject was tracked by optical sensor and investigated in this study. All subjects exhibit right hand dominance following the Edinburgh-handedness inventory~\cite{oldfield1971assessment}. No ethic approval was required.

The movement of distal arm for each subject was recorded during wrist FE motion through the Leap Motion sensor. The Leap Motion sensor offers novel solutions to track the motion of human hands and fingers and record discrete positions and gestures of distal arm within intimate proximity of operating space~\cite{iosa2015leap,yu2016fusion,xu2015bilateral}. Study addressing the tracking accuracy and robustness of the optical device has reported that an overall average accuracy of 0.7 mm was achieved among static and dynamic measurement. Deviation between desired and measured positions was less than 0.2 mm, independent of axis under experimental condition~\cite{weichert2013analysis}. Another study revealed that standard deviation between 0.0081 mm and 0.49 mm was obtained~\cite{guna2014analysis}. It indicated that the Leap Motion sensor is a reasonably precise and reliable tracking system dedicated to detecting the motion of human hands.

\begin{figure}[t]
	\centering	
	\includegraphics[height=0.55\columnwidth]{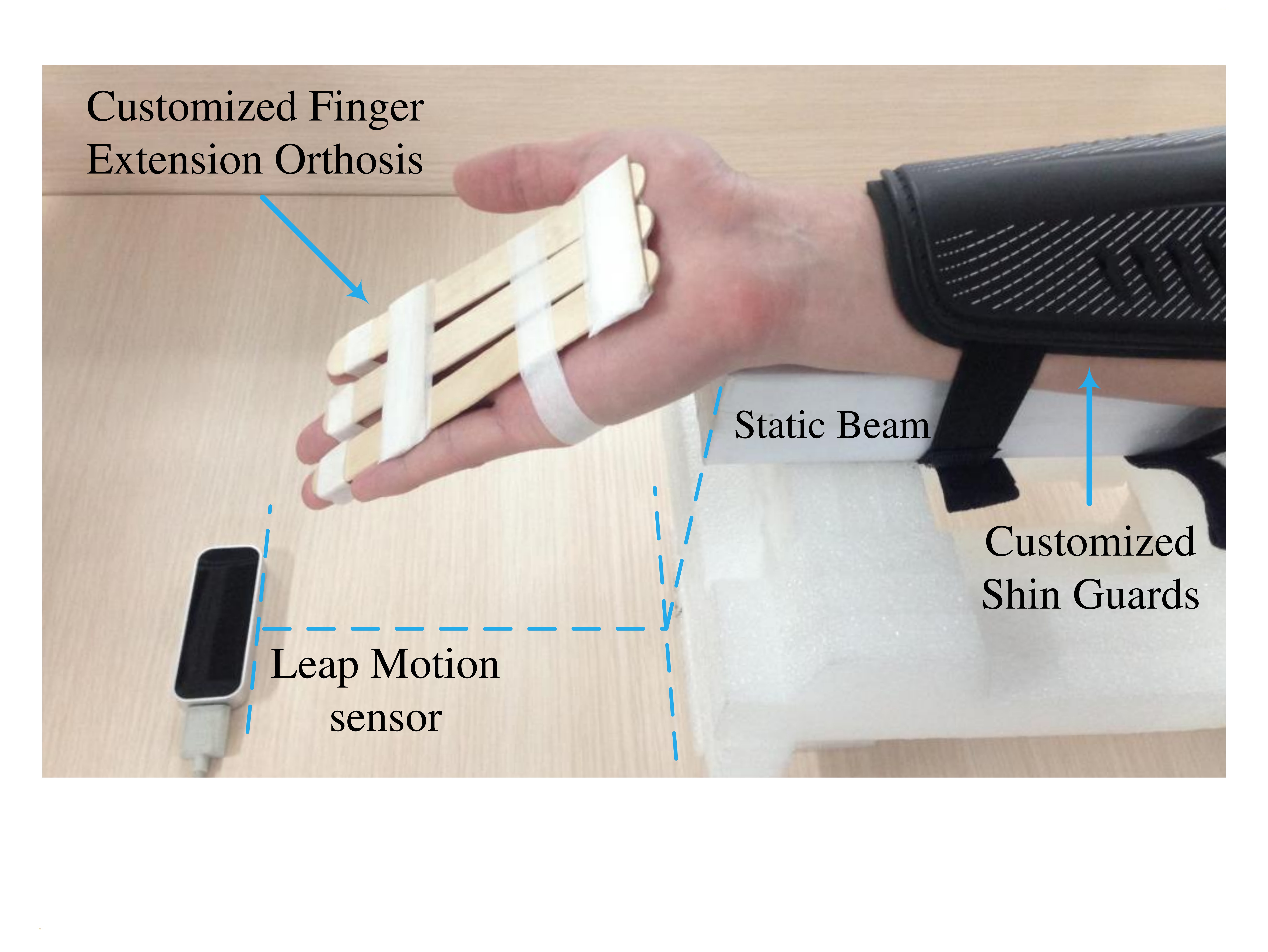}
	\caption{Experiment apparatuses and configuration including customized finger extension orthosis and shin guards which constrain undesired movement in the hand and forearm.}
	\label{setup}
\end{figure}

As illustrated in Fig.~\ref{setup}, all volunteers were seated beside the experiment table and placed the forearms onto the axially static beam, paralleling to the long axis of the beam with the elbow joint flexed about ${90^ \circ }$ and the shoulder joint abducted about ${45^ \circ }$. The forearms were constrained by customized shin guards and constantly fixed at the static beam, resulting no pronation-supination~(PS) motion in forearms. Customized finger extension orthosis was fixed on subjects' palm, constraining the relative movement in interphalangeal joints, metacarpophalangeal joints and carpometacarpal joints of the index, middle and ring fingers which were extended and coincident within volar plane of the hand. The middle finger was specifically fixed distally along the radial long axis, ensuring that the middle finger is collinear with the ${Y}$ axis of capitate which is defined according to ISB recommendation~\cite{wu2005isb}. The functional neutral position of the wrist relative to the radius was defined according to ISB recommendation and the neutral of the forearm was defined as in neutral PS position. In addition to neutral position, subjects were instructed to rotate the wrists smoothly within range from ${30^ \circ }$ of flexion to ${10^ \circ }$ of extension. The trajectories of tracked hands were illustrated by the visualizer application simultaneously. Subjects were instructed to complete 10 cycles in FE motion slowly within 40 seconds ensuring accurate tracking. One typical cycle movement included moving from neutral to extension, then to flexion and back to neutral finally.

\section{Results}

\begin{figure}[b]
	\centering
	\includegraphics[height=0.8\columnwidth]{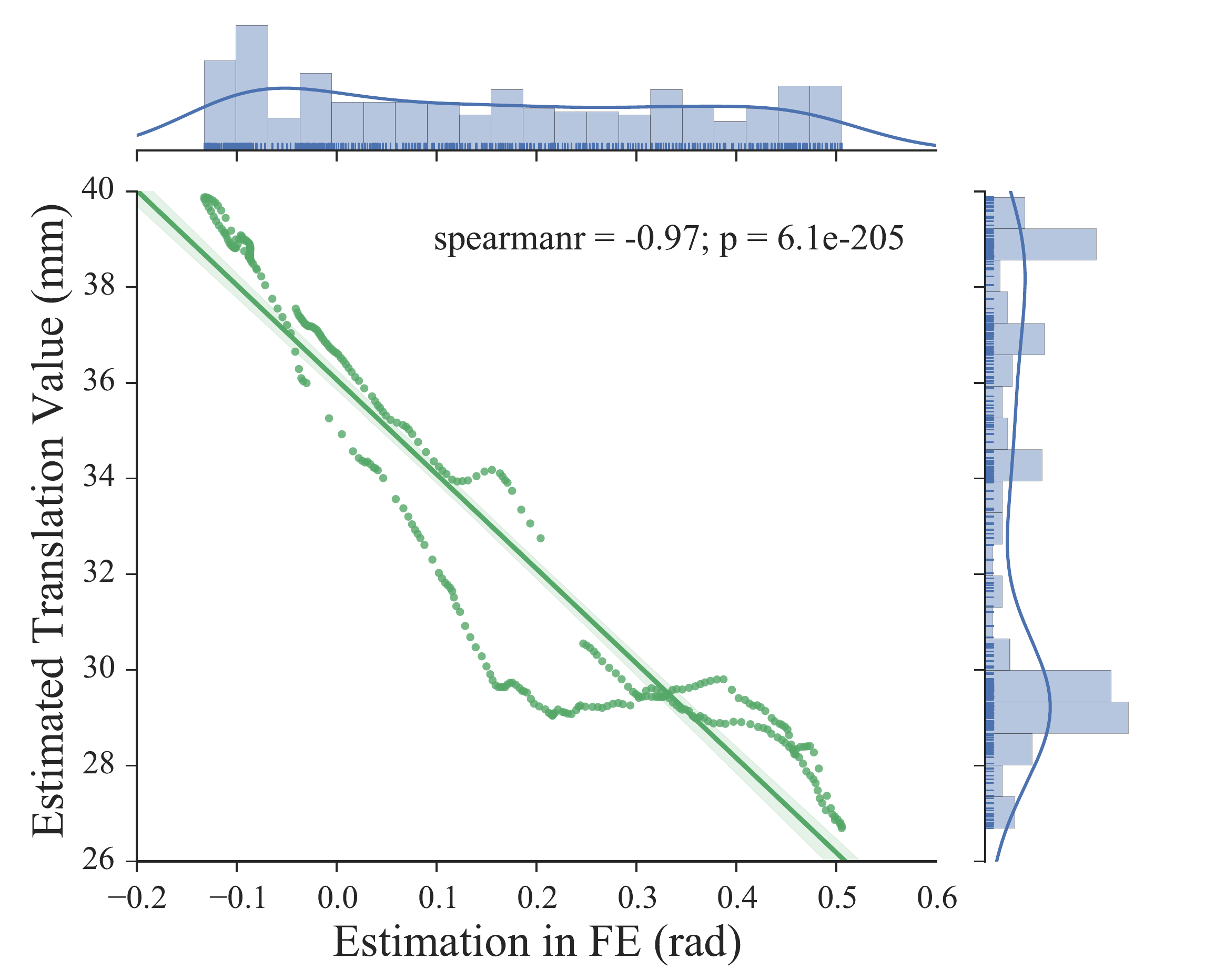}
	\caption{Calculation of measurable translation parameter ${d_2}$ in one cycle FE movement.} \label{d2fit}
\end{figure}

Experimental data were extracted and illustrated to validate the feasibility and accuracy of the improved model. As shown in Fig.~\ref{d2fit}, the translation parameter ${d_2}$ indicates that the location of wrist rotation axes was not fixed in single point. The value of ${d_2}$ changes evidently during one cycle in FE motion. Compared with the angle in FE, a more distal location of the rotation axes illustrated by larger ${d_2}$ was noted during wrist extension than that location during wrist flexion. A linear regression between ${d_2}$ and ${\beta _4}$ in FE was conducted. The Spearman's rank correlation coefficient of -0.97 ($p = 6.1{e^{ - 205}}$) indicated that statically high correlation between ${d_2}$ and ${\beta _4}$ was exhibited and the negative correlation showed that ${d_2}$ tends to decrease when ${\beta _4}$ increases. It can be concluded that carpal joints can not be modeled as simple joints and carpal rotation axes move distally during wrist extension and move proximally during wrist flexion. These results are consistent with previous biomechanical study~\cite{wolfe2000vivo}.

\begin{figure}[tb]
	\centering
	\includegraphics[height=0.75\columnwidth]{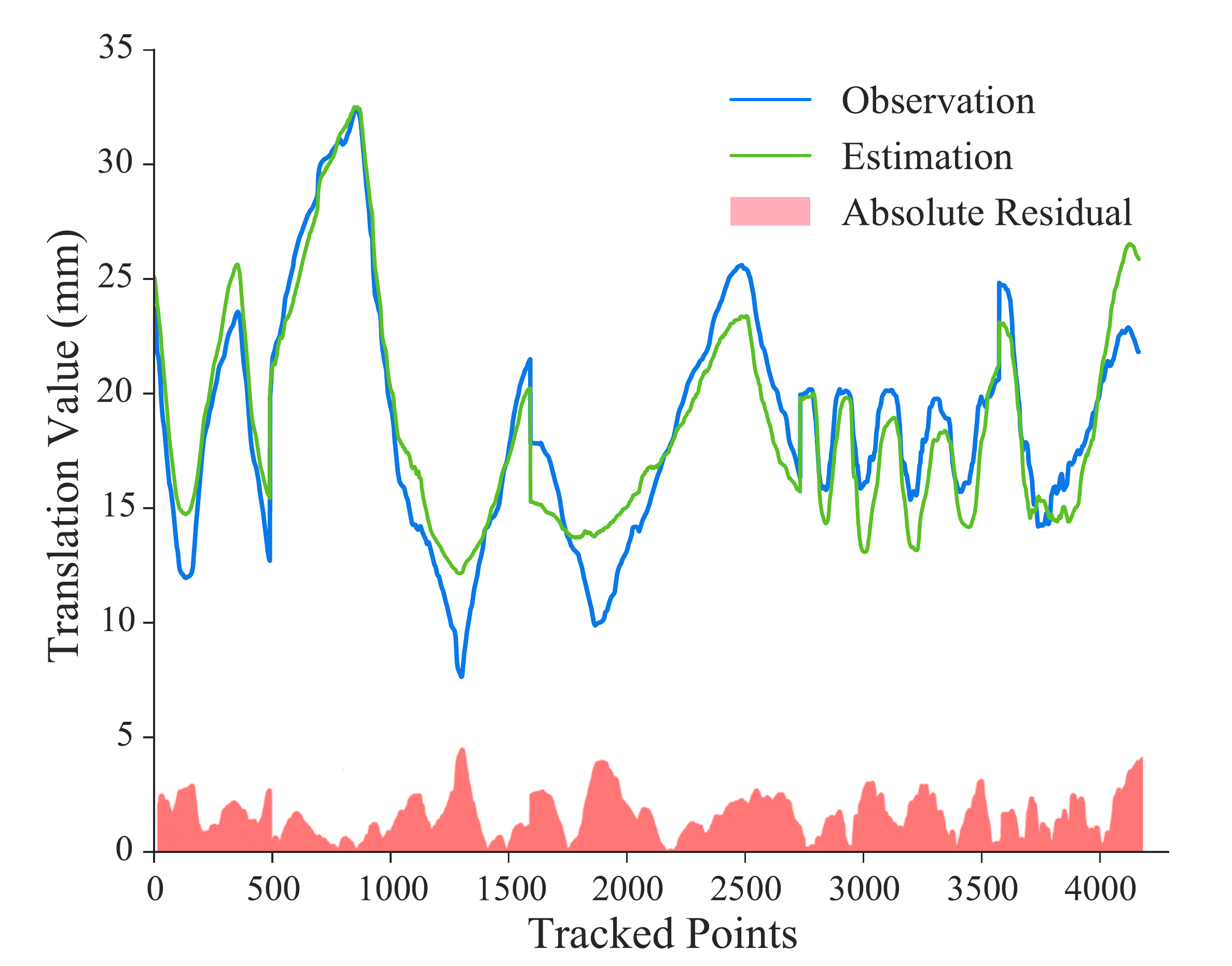}
	\caption{Results of nonlinear regression between translation parameter ${d_2}$ and estimated wrist angles ${\beta _{3,4}}$ based on nine subjects' tracking data.} \label{fit}
\end{figure}

\begin{figure}[b]
	\centering
	\includegraphics[height=0.75\columnwidth]{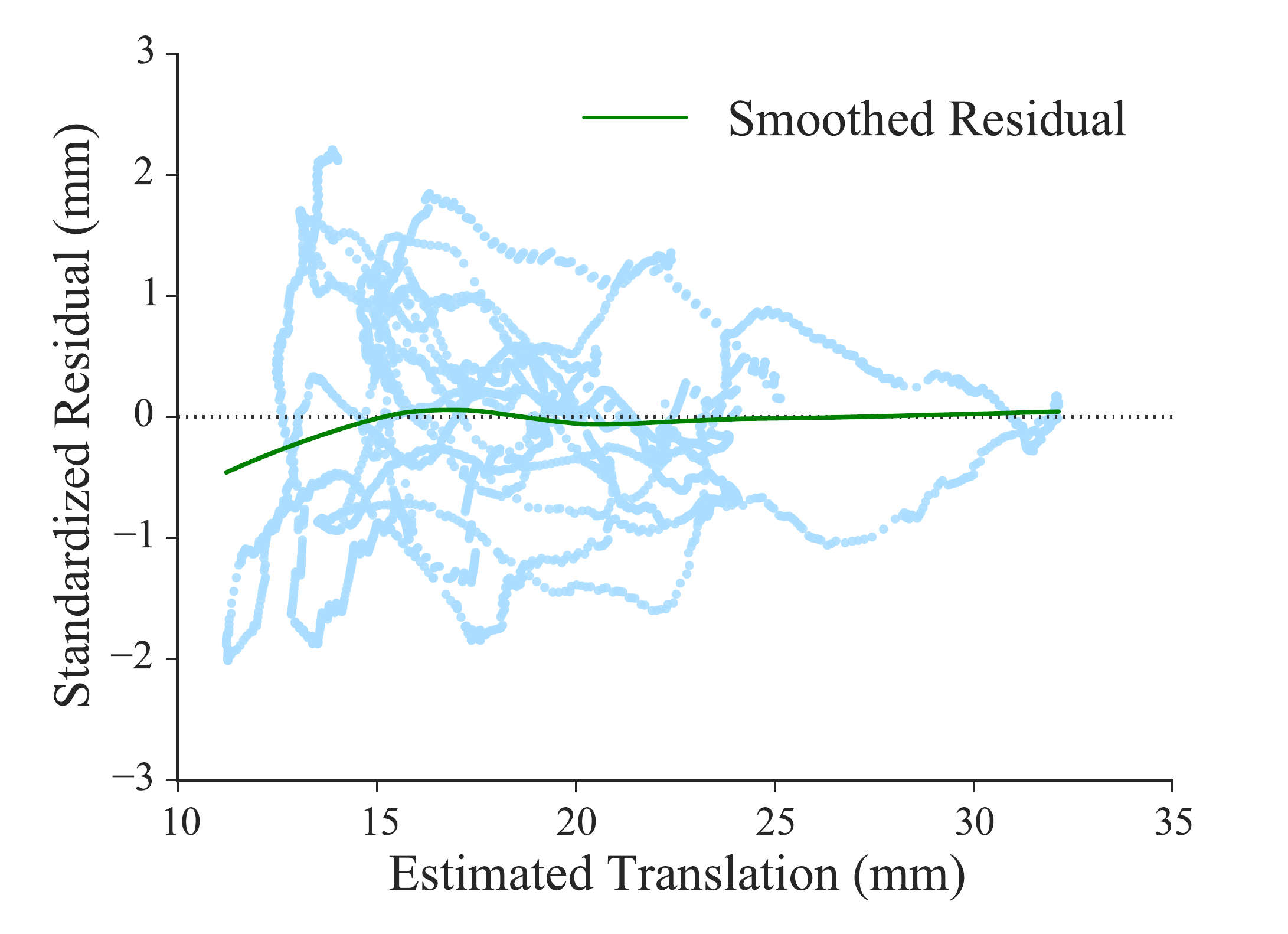}
	\caption{Interpretation of regression performance with smoothed standardized residuals.} \label{residual}
\end{figure}

Tracking data of nine subjects (two females, age $22.0 \pm 1.4$ and seven males, age $25.0 \pm 1.4$) were chosen randomly for model parameter fitting process. Data of ${\beta _{3,4}}$ derived from inverse kinematics were grouped together to fit the nonlinear function for accurate estimation of ${d_2}$, as described in~(\ref{fit}). Unknown parameters ${a_j}$ ($j = 1, \ldots ,10$) were optimized through SGA and the fitting results are illustrated in Fig.~\ref{fit} and described as
\begin{equation}\label{fit_d}
\begin{array}{l}
{{\hat d}_2} = \\
\frac{{18.00 - 290.93{\beta _3} - 29.46{\beta _4} + 2563.09{\beta _3}^2 + 37.01{\beta _4}^2 - 606.53{\beta _3}{\beta _4}{\mkern 1mu} }}{{1 - 10.60{\beta _3} - 2.23{\beta _4} + 94.62{\beta _3}^2 + 2.12{\beta _4}^2 - 25.75{\beta _3}{\beta _4}{\mkern 1mu} }}
\end{array}
\end{equation}

The standardized residual ${r_i}$ was calculated and analyzed by locally weighted scatterplot smoothing (LOWESS). As shown in Fig.~\ref{residual}, the residual plot is nearly symmetrically distributed and the scatter points are clustered around the line where ${r_i} = 0$. Besides ${r_i}$, values of SSE, RMSE, correlation coefficient ($R$) and ${R^2}$ are determined as 7565.68, 1.35, 0.96 and 0.93 respectively. All these mentioned statistical results justified the accuracy of the fitting process and illustrate quantitative connections between rotation axes and coupled wrist angles.

\begin{figure}[t]
	\centering
	\includegraphics[height=0.8\columnwidth]{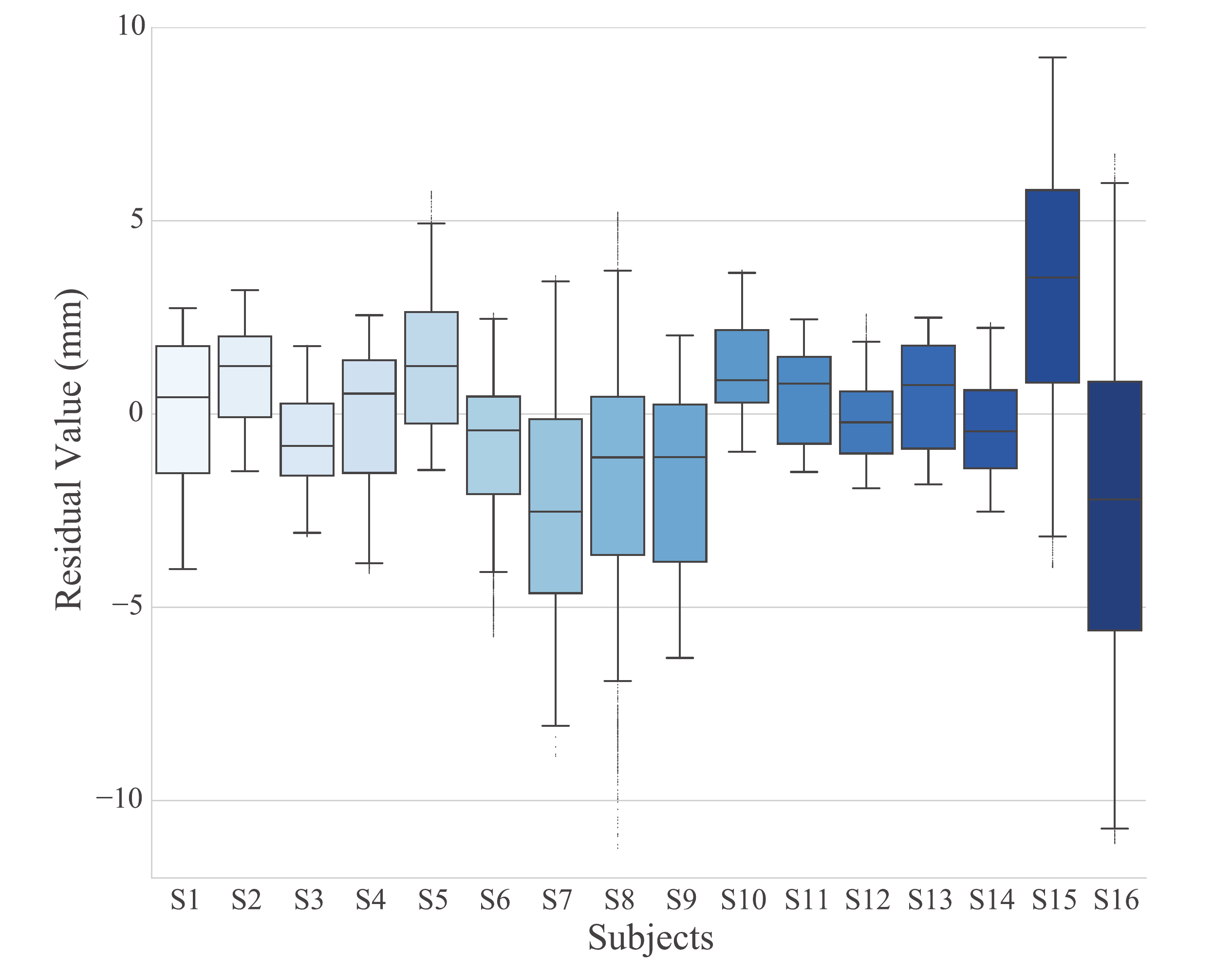}
	\caption{Residuals derived from calculated ${d_2}$ and estimated ${{\hat d}_2}$ among sixteen subjects in box plots.} \label{subjects}
\end{figure}

Tracking data of the left sixteen subjects~(2 females, age $23.5 \pm 0.7$ and 14 males, age $24.8 \pm 1.5$) were prepared to verify the fitted equation~(\ref{fit_d}). The wrist angle ${\beta _{3,4}}$ for each subjects were calculated from inverse kinematics of the improved model and were used to estimate ${d_2}$ through equation~(\ref{fit_d}). The estimation values ${{\hat d}_2}$ were taken in comparison with the calculation of ${d_2}$ derived from equation~(\ref{d_2}) to illustrate the accuracy of improved model among sixteen subjects. As illustrated in Fig.~\ref{subjects}, overall residual values among subjects fell in the range from -12 mm to 10 mm. The overall average deviation between ${{\hat d}_2}$ and ${d_2}$ across all subjects was $-0.39 \pm 3.14$ mm and average percentage error across all subjects was $8.00 \pm 3.27\% $. It can be concluded that the proposed equation estimated the location of unfixed rotation axes accurately and the results were consistent among twenty subjects statistically.

\section{DISCUSSION}

\subsection{Consistency of Wrist Biomechanics}

Experiments recruited 25 healthy young subjects without distal arm pathology, including 4 females and  21 males. Previous study has presented high uniformity between carpal kinematics of uninjured subjects. Differences of \emph{in vivo} carpal behavior and distal arm biomechanism among subjects are statistically negligible and differences between right and left wrists are not significant~\cite{wolfe2000vivo}. In addition, differences in location of rotation axes demonstrated between males and females are related to subject anatomical size~\cite{coburn2007coordinate}. It is suggested that wrist motion parameters are independent from gender and the locations of floating axes are related to carpal bones volume and the neutral centroid location~\cite{neu2001vivo,wolfe2000vivo}. Therefore, based on previous evidence addressing consistent individual carpal kinematics, it is rational to conduct confirmatory experiments according to the stated participant recruitment criteria.

\subsection{Measurement of Carpal Kinematics}

Evident differences between \emph{in vivo} and \emph{in vitro} experimental techniques are reported from previous biomechanical studies. For \emph{In vitro} studies, reflective pins and markers are implanted into carpal bones~\cite{patterson1998high,neu2000kinematic}. This invasive implant techniques may alter kinematic characteristics of overlying tendons and surrounding soft tissue. Exterior constrains imposed by kinematic markers would interfere individual carpal bones motion and mechanical interaction within carpus. Normal muscular contraction and tendon loading are simulated by attached mechanism where uncertain deviation from \emph{in vivo} simulation would occur inevitably and minute changes in muscular dynamics may affect wrist motion significantly~\cite{patterson1998high}. Computed tomography~(CT) scanning is commonly adopted by \emph{in vivo} studies and issues addressing safe radiation exposure may limit the duration of scanning protocol and the number of experimental samples~\cite{coburn2007coordinate,yu2011fmri}. Compared with implanted measurement and scanning techniques, the proposed noninvasive markerless method offers more applicable distal arm optical tracking solution for human-robot interaction and restores natural carpus motion authentically. In addition, statistically negligible differences were reported between static and dynamic wrist motion and the tracking data derived from static scanning procedure approximates dynamic passive carpal kinematics. It can be concluded that carpal bone kinematics can be measured either statically or dynamically~\cite{foumani2009vivo}. Therefore, it is rational to conduct the proposed optical tracking procedure during relatively slow movement pace.

\subsection{Implementation Rationale for Robotic Neurorehabilitation}

Functional assessment for neurorehabilitation and mechanical implementation for robotic rehabilitation devices are fundamentally based on biomenchanically rational kinematic model of human body~\cite{rinderknecht2018reliability,nef2009armin}. Under the assumption that only rotation motions exist in decoupled carpal joints, conventional kinematic model of the upper limb fails to estimate the posture of end-effector and wrist joint precisely. This inevitable limitation and undesired misalignment between joint kinematics and robotic evaluation will bring unreliable parameters for rehabilitation assessment in joint space. The proposed improved kinematic model of wrist motion has justified the existence of measurable unfixed axes in carpal rotation and provided accurate estimation of axes location based on coupled wrist angles. Therefore, kinematic assessment of distal arms in joint space is feasible and effective based on the accurate model, providing objective and quantitative measurement for neurorehabilitation. In addition, D-H parameter notation is a generic and standardized robotic notation~\cite{santos2006reported}. The proposed kinematic model of wrist motion can be directly adopted in robotic fields and is essential for robotic implementation of exoskeleton and robotic neurorehabilitation evaluation.

\section{CONCLUSIONS}

Accurate evaluation of upper extremity kinematics are keys for better understanding of physical human-robot interactions to deliver robotic neurorehabilitation and assistance. In specific, realistic biomechanical modeling of human wrist is crucial for ergonomic designs of assistive exoskeletons and human-robot collaboration. In this study, the proposed improved kinematic model has justified the existence of measurable unfixed axes in carpal rotation which supports prior studies indicate that inevitable misalignment and oversimplification between robotic representation and human joints will occur. The accurate estimation of axes location are achieved through coupled wrist angles with nonlinear regression. Experiments with uninjured subjects have validated the improved model through optical tracking method and numerical optimization based on robotic modeling. Therefore, kinematic assessment of distal arms in joint space is feasible and effective based on this proposed model, enabling quantitative implementation for physical human-robot interaction and robotic neurorehabilitation.









\bibliographystyle{IEEEtran}
\bibliography{changral}

\end{document}